\documentclass{article}
\usepackage{booktabs}
\usepackage{tabularx}
\usepackage{array}
\usepackage{makecell}
\usepackage{spconf,amsmath,graphicx,hyperref}
\usepackage{amsmath,amsfonts}
\usepackage{subfiles} 
\usepackage{algorithm}
\usepackage{array}
\usepackage[caption=false,font=normalsize,labelfont=sf,textfont=sf]{subfig}
\usepackage{cite}
\usepackage{textcomp}
\usepackage{stfloats}
\usepackage{url}
\usepackage{verbatim}
\usepackage{graphicx}
\usepackage{cite}
\usepackage{booktabs}
\usepackage{algorithmicx}
\usepackage{algpseudocode}
\usepackage{mathrsfs}
\usepackage{amsmath} 
\usepackage{multirow}
\usepackage{color}
\usepackage{tabularx}
\usepackage[dvipsnames]{xcolor}
\floatname{algorithm}{Algorithm} 
\usepackage{lineno}
\usepackage[table]{xcolor}
\usepackage{amsmath}
\usepackage{fontawesome5}
\usepackage{xcolor}
\usepackage{siunitx}
\usepackage{tabularx}
\usepackage{subcaption}
\usepackage{array}

\definecolor{gold}{RGB}{212,175,55}
\definecolor{silver}{RGB}{192,192,192}
\definecolor{bronze}{RGB}{205,127,50}

\usepackage{booktabs}
\usepackage{multirow}
\usepackage{caption}
\usepackage{array}
\usepackage{amsmath}
\usepackage{algorithm}
\usepackage{algpseudocode}

\usepackage{graphicx}
\usepackage[table,xcdraw]{xcolor}  
\usepackage{longtable}

\usepackage{booktabs}
\usepackage{tabularx}
\usepackage{array}
\usepackage{makecell}
\usepackage{siunitx}
\usepackage[table]{xcolor}
\usepackage{colortbl}

\newcolumntype{L}{>{\raggedright\arraybackslash}X}

\definecolor{TopOne}{HTML}{E1E1E1}
\definecolor{TopTwo}{HTML}{ECECEC}
\definecolor{TopThree}{HTML}{F6F6F6}

\definecolor{BarOne}{HTML}{8E8E8E}
\definecolor{BarTwo}{HTML}{A8A8A8}
\definecolor{BarThree}{HTML}{C0C0C0}

\definecolor{Divider}{HTML}{D8D8D8}
\usepackage{booktabs}
\usepackage{multirow}

\title{Towards Understanding LLM-Based Log Anomaly Detection: An Empirical Study of Performance, Efficiency, and Robustness}
\name{
Bin Li$^{1*}$,
Dongdong Wang$^{2*}$,
Siyang Lu$^{3\dagger}$
\thanks{$^{*}$Equal contribution. $^{\dagger}$Corresponding author.}
}

\address{
$^{1}$School of Cyberspace Science and Technology, Beijing Jiaotong University, Beijing, China\\
$^{2}$College of Design, Construction, and Planning, University of Florida, Gainesville, FL, USA\\
$^{3}$School of Computer and Technology, Beijing Jiaotong University, Beijing, China\\
24120434@bjtu.edu.cn, dongdongwang@ufl.edu, sylu@bjtu.edu.cn
}

\begin{document}
%
\maketitle

\begin{abstract}
Large language models (LLMs) have demonstrated promising performance in log anomaly detection, yet how their adaptation strategies, architectures, and deployment configurations affect detection effectiveness remains insufficiently understood. To investigate these factors, we conduct a systematic empirical analysis across three public log datasets, examining different adaptation strategies, model architectures, parameter scales, and quantization settings. Our results reveal substantial performance differences across adaptation strategies, while model scaling yields varying detection gains across datasets. We further observe that models with comparable detection accuracy can exhibit markedly different computational costs, and that low-bit quantization largely preserves detection performance in the evaluated configurations. Finally, we examine detection robustness under structural, semantic, and label noise at different perturbation levels. These findings provide empirical insights into the performance, efficiency, and robustness of LLM-based log anomaly detection, highlighting practical considerations beyond conventional accuracy-oriented evaluation.

\end{abstract}


\section{INTRODUCTION}
\label{sec:introduction}

System logs record the runtime states and execution behaviors of modern computing systems and have been widely used for fault diagnosis and anomaly detection. Traditional log anomaly detection methods~\cite{du2017deeplog, hou2026log, tan2026gtcl, lu2023ssdlog, lu2018detecting} typically model parsed event templates or their sequential patterns, but their reliance on log parsing and discrete event representations may introduce parsing errors and lose semantic information. With the rapid development of large language models (LLMs) ~\cite{zhao2026survey, naveed2025comprehensive}, recent studies have begun to exploit their strong semantic modeling capabilities to capture richer semantic and contextual information from textual logs. Various adaptation strategies, including prompting and parameter-efficient fine-tuning~\cite{naveed2025comprehensive, zhang2025parameter}, have further demonstrated the potential of LLMs for log anomaly detection~\cite{he2024llmelog, guan2024logllm}.

However, current evaluations of LLM-based log anomaly detectors are largely dominated by predictive metrics such as Precision, Recall, and F1-score, providing only a partial view of practical performance. Higher accuracy may come at the cost of longer inference time, greater adaptation overhead, or a larger deployment footprint, while quantization and noisy logs may further affect reliability. Thus, accuracy alone cannot fully characterize the practical trade-offs of LLM-based log anomaly detection.

To address this limitation, we propose a six-dimensional evaluation framework covering Detection Performance, Inference Efficiency, Adaptation Efficiency, Deployment Efficiency, Quantization Resilience, and Noise Robustness. We benchmark representative adaptation strategies, model architectures, model scales, and quantization settings across BGL, HDFS, and Thunderbird to reveal trade-offs overlooked by accuracy-oriented evaluation.

The main contributions of this work are summarized as follows:
\begin{itemize}
    \item \textbf{Detection performance analysis.} We systematically investigate the effects of adaptation strategies, model architectures, and parameter scales on LLM-based log anomaly detection across three public datasets, revealing performance differences across adaptation strategies and varying detection gains from model scaling.
    \item \textbf{Efficiency analysis.} We examine the trade-offs among detection performance, inference efficiency, adaptation overhead, and deployment cost, showing that comparable detection accuracy can entail different computational costs.
    \item \textbf{Robustness analysis.} We evaluate the effects of low-bit quantization and structural, semantic, and label noise, revealing largely preserved detection performance under quantization and varying sensitivity to noise.
\end{itemize}

\begin{figure}
 \includegraphics[width=\columnwidth]{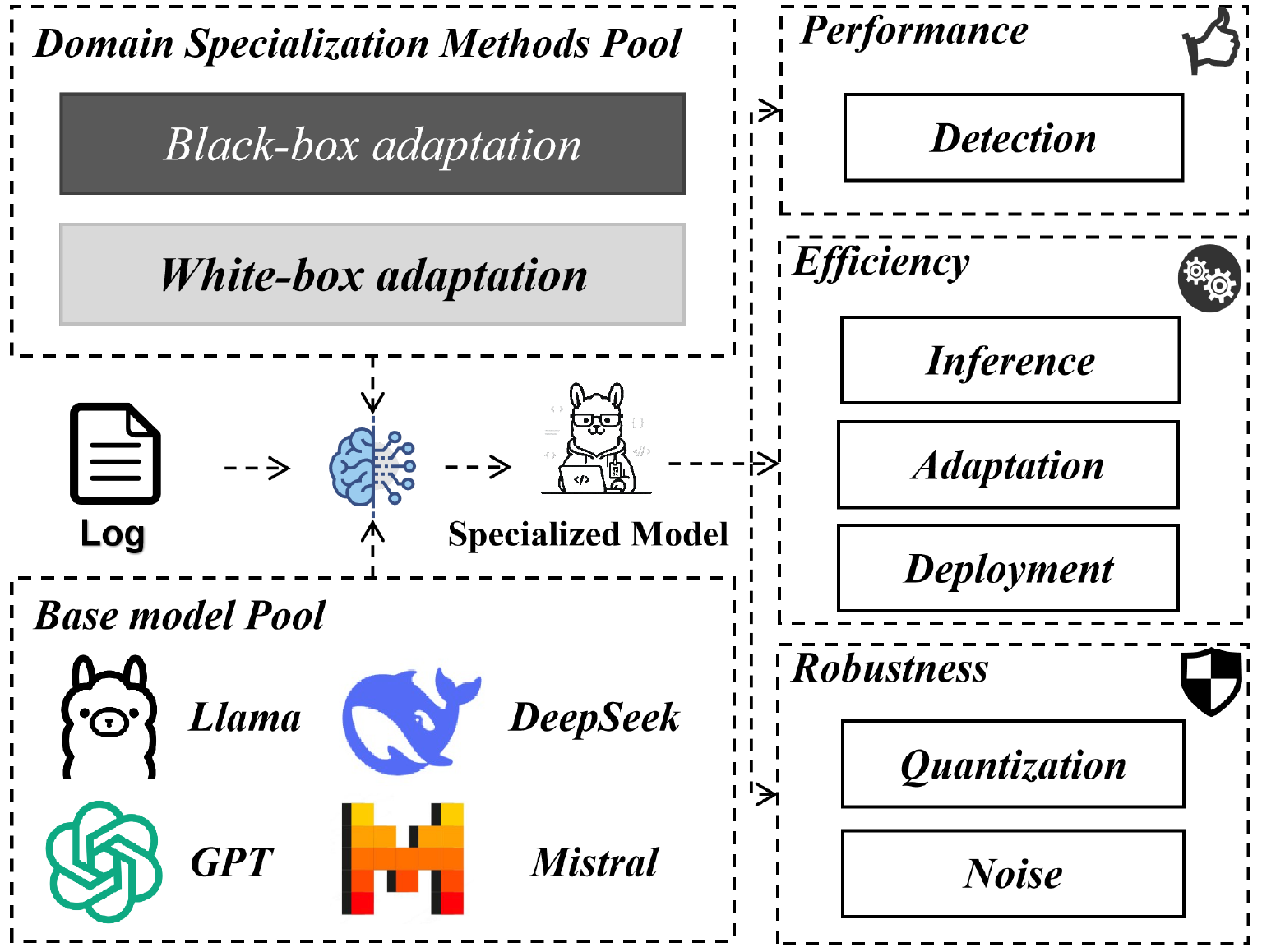}
    \caption{Overview of the proposed evaluation framework. Specialized language models are evaluated across six dimensions covering detection performance, inference efficiency, adaptation efficiency, deployment efficiency, quantization resilience, and noise robustness.}
   \label{fig:eval}
\end{figure}

\begin{table}[h]
\centering
\caption{Configurations of evaluated language models.}
\label{tab:baseline}
\scriptsize
\renewcommand{\arraystretch}{0.85}
\begin{tabular*}{\columnwidth}{@{\extracolsep{\fill}}lcr@{}}
\toprule
\textbf{Model} & \textbf{Architecture} & \textbf{Context} \\
\midrule
Llama2-7B & 32 Decoder & 4096 \\
Llama3-8B & 32 Decoder & 8192 \\
RoBERTa~\cite{liu2019roberta} & 24 Encoder & 512 \\
DeBERTa~\cite{hedeberta} & 24 Encoder & 512 \\
BART~\cite{lewis2020bart} & 12 Enc. + 12 Dec. & 1024 \\
T5~\cite{raffel2020exploring} & 24 Enc. + 24 Dec. & 512 \\
DeepSeek-7B~\cite{guo2025deepseek} & 28 Decoder & 131072 \\
GPT-2 Variants & 24/36/48 Decoder & 1024 \\
\bottomrule
\end{tabular*}
\vspace{-2mm}
\end{table}

\begin{table*}[t]
\centering
\caption{Detection performance (\%) across BGL, HDFS, and Thunderbird.}
\label{tab:performance}
\scriptsize
\renewcommand{\arraystretch}{0.85}
\setlength{\tabcolsep}{0pt}

\begin{tabular*}{\textwidth}{@{\extracolsep{\fill}}llcccccccccccc@{}}
\toprule
\multirow{2}{*}{\textbf{Method}} &
\multirow{2}{*}{\textbf{Setting}} &
\multicolumn{3}{c}{\textbf{BGL}} &
\multicolumn{3}{c}{\textbf{HDFS}} &
\multicolumn{3}{c}{\textbf{Thunderbird}} &
\multicolumn{3}{c}{\textbf{Average}} \\
\cmidrule(lr){3-5}\cmidrule(lr){6-8}\cmidrule(lr){9-11}\cmidrule(l){12-14}
& & P & R & F1 & P & R & F1 & P & R & F1 & P & R & F1 \\
\midrule

Head-only        & White-box & 96.32 & 81.03 & 88.02 & 99.87 & 52.77 & 69.05 & 96.39 & 99.95 & 98.13 & 97.53 & 77.92 & 85.07 \\
Layer-Norm       & White-box & 95.76 & 83.53 & 89.23 & 100.00 & 37.23 & 54.26 & 96.39 & 99.95 & 98.13 & 97.38 & 73.57 & 80.54 \\
Frozen           & White-box & 99.93 & 99.70 & 99.82 & 99.50 & 99.13 & 99.32 & 96.50 & 99.73 & 98.09 & 98.64 & 99.52 & \textbf{99.08} \\
LoRA             & White-box & 99.50 & 99.67 & 99.58 & 99.46 & 98.43 & 98.94 & 96.44 & 99.95 & 98.16 & 98.47 & 99.35 & 98.89 \\
Prefix           & White-box & 80.40 & 71.50 & 75.69 & 92.71 & 83.03 & 87.60 & 79.95 & 100.00 & 88.86 & 84.35 & 84.84 & 84.05 \\
Adapter-par.     & White-box & 99.90 & 99.37 & 99.63 & 99.37 & 89.73 & 94.31 & 92.13 & 98.81 & 95.36 & 97.13 & 95.97 & 96.43 \\
Adapter-ser.     & White-box & 100.00 & 99.53 & 99.77 & 99.38 & 90.50 & 94.73 & 91.09 & 100.00 & 95.34 & 96.82 & 96.68 & 96.61 \\
\midrule

Zero-shot        & Prompt-based & 78.09 & 49.90 & 60.89 & 51.35 & 85.70 & 64.22 & 65.39 & 37.40 & 47.58 & 64.94 & 57.67 & 57.56 \\
Few-shot         & Prompt-based & 54.70 & 86.50 & 67.02 & 56.80 & 83.30 & 67.54 & 75.14 & 32.03 & 44.92 & 62.21 & 67.28 & 59.83 \\
CoT              & Prompt-based & 56.03 & 78.13 & 65.26 & 59.49 & 71.43 & 64.92 & 76.14 & 40.17 & 52.60 & 63.89 & 63.24 & 60.93 \\
Prompt-Tuning    & Prompt-based & 99.53 & 99.00 & 99.27 & 99.50 & 98.87 & 99.18 & 96.48 & 99.68 & 98.05 & 98.50 & 99.18 & \textbf{98.83} \\
P-Tuning v1      & Prompt-based & 83.76 & 71.00 & 76.85 & 80.67 & 98.50 & 88.70 & 95.99 & 99.97 & 97.94 & 86.81 & 89.82 & 87.83 \\
\midrule

RoBERTa          & Architecture & 99.93 & 98.47 & 99.19 & 99.51 & 95.53 & 97.48 & 96.53 & 99.78 & 98.13 & 98.66 & 97.93 & 98.27 \\
DeBERTa          & Architecture & 99.53 & 99.00 & 99.26 & 99.45 & 95.70 & 97.54 & 96.50 & 99.68 & 98.06 & 98.49 & 98.13 & 98.29 \\
BART             & Architecture & 99.19 & 97.53 & 98.35 & 97.86 & 95.87 & 96.85 & 96.39 & 99.95 & 98.13 & 97.81 & 97.78 & 97.78 \\
T5               & Architecture & 99.44 & 95.00 & 97.17 & 98.56 & 86.43 & 92.10 & 96.26 & 99.81 & 98.00 & 98.09 & 93.75 & 95.76 \\
DeepSeek-7B      & Architecture & 99.97 & 99.67 & 99.82 & 99.47 & 99.73 & 99.60 & 96.53 & 99.68 & 98.08 & 98.66 & 99.69 & \textbf{99.17} \\
Llama3-8B        & Architecture & 99.87 & 99.70 & 99.78 & 99.44 & 99.67 & 99.55 & 96.48 & 99.68 & 98.05 & 98.60 & 99.68 & 99.13 \\
\midrule

Llama2-7B (FP16) & Quantization & 99.50 & 99.67 & 99.58 & 99.46 & 98.43 & 98.94 & 96.44 & 99.95 & 98.16 & 98.47 & 99.35 & 98.89 \\
Llama2-7B (INT8) & Quantization & 99.89 & 99.70 & 99.80 & 99.50 & 99.10 & 99.30 & 96.14 & 100.00 & 98.03 & 98.51 & 99.60 & \textbf{99.04} \\
Llama2-7B (INT4) & Quantization & 99.93 & 99.80 & 99.87 & 99.49 & 98.43 & 98.96 & 96.43 & 99.84 & 98.11 & 98.62 & 99.36 & 98.98 \\
\midrule

GPT-2 Medium     & Model scale & 99.69 & 96.90 & 98.28 & 99.93 & 45.57 & 62.59 & 95.71 & 5.42 & 10.26 & 98.44 & 49.30 & 57.04 \\
GPT-2 Large      & Model scale & 99.52 & 97.20 & 98.35 & 99.30 & 66.27 & 79.49 & 96.42 & 99.35 & 97.86 & 98.41 & 87.61 & 91.90 \\
GPT-2 XL         & Model scale & 99.72 & 95.60 & 97.62 & 99.42 & 80.07 & 88.70 & 96.30 & 99.76 & 98.00 & 98.48 & 91.81 & \textbf{94.77} \\

\bottomrule
\end{tabular*}
\vspace{-2mm}
\end{table*}

\sisetup{
  round-mode=places,
  round-precision=2
}

\begin{table}[t]
\centering
\caption{Comparison of detection performance, efficiency, and robustness across representative configurations. All metrics are reported on a 0--100 scale (higher is better).}
\label{tab:efficiency}
\scriptsize
\setlength{\tabcolsep}{2.2pt}
\renewcommand{\arraystretch}{0.85}

\begin{tabular*}{\columnwidth}{@{\extracolsep{\fill}}llcccccc@{}}
\toprule
\multirow{2}{*}{\textbf{Method}} &
\multirow{2}{*}{\textbf{Setting}} &
\multicolumn{1}{c}{\textbf{Perf.}} &
\multicolumn{3}{c}{\textbf{Efficiency}} &
\multicolumn{2}{c}{\textbf{Robustness}} \\
\cmidrule(lr){3-3}\cmidrule(lr){4-6}\cmidrule(l){7-8}
& & Det. & Inf. & Adapt. & Deploy. & Quant.R. & Noise \\
\midrule
DeBERTa       & Arch.  & 98.29 & 96.06 & 94.20 & 97.37 & 99.98 & 99.74 \\
BART          & Arch.  & 97.78 & 99.58 & 85.53 & 97.37 & 99.92 & 91.01 \\
DeepSeek-7B   & Arch.  & 99.17 & 94.83 & 93.41 & 55.05 & 99.95 & 99.78 \\
\addlinespace[1pt]
GPT-2 XL      & Scale  & 94.77 & 99.03 & 99.57 & 90.56 & 99.98 & 98.54 \\
\addlinespace[1pt]
LoRA          & WB     & 98.89 & 89.63 & 99.87 & 60.00 & 99.95 & 99.92 \\
Frozen        & WB     & 99.08 & 89.49 & 93.87 & 60.00 & 98.67 & 99.78 \\
Adapter-par.  & WB     & 96.43 & 90.97 & 97.99 & 60.00 & 100.00 & 100.00 \\
\addlinespace[1pt]
Prompt-Tuning & Prompt & 98.83 & 89.30 & 99.99 & 60.00 & 99.78 & 99.69 \\
CoT           & Prompt & 60.93 & 91.96 & 100.00 & 60.00 & 97.34 & 100.00 \\
\addlinespace[1pt]
LoRA INT8     & Quant. & 99.04 & 95.53 & 99.87 & 79.13 & 99.72 & 99.95 \\
\bottomrule
\end{tabular*}

\vspace{1pt}
\parbox{\columnwidth}{\scriptsize
\textit{Note:} Arch.: architecture; WB: white-box adaptation;
Quant.: quantization; Det.: detection; Inf.: inference;
Adapt.: adaptation; Deploy.: deployment; Quant.R.: quantization resilience.}
\vspace{-2mm}
\end{table}





\section{Experimental Design}
\label{sec:experimental_design}

We conduct a systematic empirical analysis of LLM-based log anomaly detection across three public datasets. Our experiments examine the effects of adaptation strategies, model architectures, parameter scales, and quantization settings on detection performance. We further investigate computational efficiency and robustness under different noise conditions.

\subsection{Datasets and Evaluation Protocol}
\label{sec:datasets}

We conduct experiments on three widely used public log datasets: BGL~\cite{oliner2007supercomputers}, HDFS, and Thunderbird~\cite{oliner2007supercomputers}. BGL contains system logs collected from the Blue Gene/L supercomputer, with anomaly labels provided at the log-event level. HDFS records the execution behaviors of the Hadoop Distributed File System, where log sequences are grouped by block identifiers and assigned normal or anomalous labels. Thunderbird contains large-scale supercomputer logs collected at Sandia National Laboratories.

For all datasets, we preserve the original chronological order of log entries rather than randomly shuffling the entire dataset. Earlier log sequences are used for model adaptation, while later sequences are retained for evaluation. This chronological sampling strategy reflects practical deployment scenarios, where models are adapted using historical logs and subsequently evaluated on future system events, while reducing information leakage between training and test sets.

\subsection{Evaluated Models and Adaptation Strategies}
\label{sec:models}

We investigate four factors that may affect LLM-based log anomaly detection: adaptation strategy, model architecture, parameter scale, and quantization precision. The evaluated model configurations are summarized in Table~\ref{tab:baseline}.

\textbf{Adaptation strategies.} We evaluate both white-box and prompt-based adaptation approaches. White-box methods include Head-only, Layer-Norm tuning, Frozen-layer tuning, LoRA, Prefix tuning, and serial and parallel Adapters. Head-only serves as a control baseline, while the remaining methods represent different forms of parameter-efficient adaptation. Prompt-based approaches include Zero-shot, Few-shot, Chain-of-Thought (CoT), Prompt-Tuning, and P-Tuning v1, covering both hard- and soft-prompt strategies.

\textbf{Model architectures.} We compare representative encoder-only, encoder--decoder, and decoder-only models. Specifically, RoBERTa and DeBERTa represent encoder-only architectures, BART and T5 represent encoder--decoder architectures, and DeepSeek-7B and Llama3-8B represent decoder-only architectures.

\textbf{Model scale.} To investigate the relationship between parameter scale and detection performance, we compare GPT-2 Medium, Large, and XL, which belong to the same model family but differ in parameter size.

\textbf{Model quantization.} We evaluate Llama2-7B under 16-bit floating-point, 8-bit, and 4-bit configurations to examine the effects of numerical precision reduction on detection performance and deployment efficiency.

\subsection{Evaluation Metrics}
\label{sec:metrics}
We evaluate LLM-based log anomaly detection from three complementary perspectives: detection performance, computational efficiency, and robustness.

\textbf{Detection performance.} We report Precision, Recall, and F1-score, using F1-score as the primary metric. Average detection performance is measured by $\overline{F1}=|\mathcal{D}|^{-1}\sum_{d\in\mathcal{D}}F1_d$, where $\mathcal{D}$ denotes the set of evaluated datasets. Dataset-specific results are also reported to examine performance variations across log sources.

\textbf{Computational efficiency.} We assess inference, adaptation, and deployment efficiency. Average inference latency is defined as $t_{\mathrm{avg}}=N^{-1}\sum_{i=1}^{N}t_i$, where $t_i$ is the latency of the $i$-th sampled test sequence. Under the same hardware environment, inference efficiency is normalized as
\begin{equation}
E_{\mathrm{inf}}=100\left(1-\frac{t_{\mathrm{avg}}-t_{\min}}{t_{\max}-t_{\min}}\right),
\label{eq:inference_eff}
\end{equation}
where $t_{\min}$ and $t_{\max}$ are the minimum and maximum average latencies across evaluated methods, respectively. Adaptation efficiency is measured by $E_{\mathrm{adapt}}=100(1-P_{\mathrm{train}}/P_{\mathrm{base}})$, where $P_{\mathrm{train}}$ and $P_{\mathrm{base}}$ denote the numbers of trainable and total model parameters, respectively. Deployment efficiency is measured using the model-size proxy $P_{\mathrm{size}}=2P_{\mathrm{base}}/10^9$ and its normalized score:
\begin{equation}
E_{\mathrm{deploy}}=100\left[1-\tanh\left(\frac{P_{\mathrm{size}}}{t}\right)\right],
\label{eq:deployment_eff}
\end{equation}
where $t$ controls the sensitivity to model scale. Higher inference, adaptation, and deployment scores indicate greater relative efficiency.


\textbf{Robustness.} We assess quantization resilience by measuring F1-score preservation after numerical precision reduction~\cite{dettmers2023qlora,xiao2023smoothquant}, using $E_{\mathrm{quant}}=100(1-|F1_{\mathrm{float}}-F1_{\mathrm{quant}}|/F1_{\mathrm{float}})$. Noise robustness is evaluated under structural, semantic, and label noise at injection ratios of 0.05, 0.15, and 0.25, using $E_{\mathrm{noise}}=100(1-|F1_{\mathrm{clean}}-F1_{\mathrm{noisy}}|/F1_{\mathrm{clean}})$. Here, $F1_{\mathrm{float}}$ and $F1_{\mathrm{quant}}$ denote the F1-scores of full-precision and quantized models, while $F1_{\mathrm{clean}}$ and $F1_{\mathrm{noisy}}$ denote those obtained under clean and noisy conditions, respectively. For each method, we first compute the quantization and noise scores under each corresponding evaluation setting and then report their arithmetic means in Table~\ref{tab:efficiency}. Higher scores indicate greater preservation of detection performance. For hard-prompt methods (Zero-shot, Few-shot, and CoT), outputs without valid target labels (\emph{Normal} or \emph{Abnormal}) are excluded from metric computation, following the original evaluation protocol.

\section{Detection Performance Analysis}
\label{sec:performance}

Table~\ref{tab:performance} reports the detection performance of different methods on BGL, HDFS, and Thunderbird. We analyze the results from three perspectives: adaptation strategies, model architectures, and parameter scales.




\textbf{Impact of adaptation strategies.} White-box adaptation and soft-prompt methods generally outperform the evaluated hard-prompt strategies. Among white-box methods, Frozen-layer tuning and LoRA achieve average F1-scores of 99.08 and 98.89, respectively, followed by serial and parallel Adapters with 96.61 and 96.43. In contrast, Head-only, Layer-Norm, and Prefix tuning obtain 85.07, 80.54, and 84.05. Among prompt-based methods, Prompt-Tuning achieves 98.83, substantially higher than P-Tuning v1 (87.83) and the hard-prompt methods, including Zero-shot (57.56), Few-shot (59.83), and CoT (60.93). Overall, adaptation strategy has a substantial impact on detection performance, with soft prompting and several white-box approaches consistently outperforming hard prompting.



\textbf{Impact of model architectures.} Different backbone architectures achieve competitive detection performance across the three datasets, with RoBERTa and DeBERTa obtaining average F1-scores of 98.27 and 98.29, BART and T5 achieving 97.78 and 95.76, and DeepSeek-7B and Llama3-8B reaching 99.17 and 99.13, respectively. However, the differences are dataset-dependent: F1-scores are tightly clustered between 98.00 and 98.13 on Thunderbird, but range from 92.10 to 99.60 on HDFS. These results suggest that the relative performance of different model configurations varies across datasets; moreover, because the evaluated models also differ in parameter scale and pretraining characteristics, the comparisons do not isolate the effect of architecture alone.




\textbf{Impact of model scale.} We further examine the relationship between model scale and detection performance using three GPT-2 variants. GPT-2 Medium, Large, and XL achieve average F1-scores of 57.04, 91.90, and 94.77, respectively, but the gains vary substantially across datasets. On BGL, all three variants perform similarly, with F1-scores ranging from 97.62 to 98.35, whereas on HDFS the F1-score increases from 62.59 to 79.49 and 88.70 as model scale increases. The difference is more pronounced on Thunderbird, where GPT-2 Medium achieves only 10.26, compared with 97.86 and 98.00 for Large and XL. These results indicate that increasing model scale can substantially improve detection performance on some datasets, but the gains are strongly dataset-dependent rather than uniform.

\section{Efficiency and Robustness Analysis}
\label{sec:efficiency_robustness}

We also examine the computational characteristics of the evaluated models and their sensitivity to reduced numerical precision and noisy log data. These analyses provide complementary insights into practical considerations that are not captured by detection metrics alone.

\textbf{Detection performance vs. computational efficiency.} Models with comparable detection performance can exhibit different computational characteristics. As shown in Tables~\ref{tab:performance} and~\ref{tab:efficiency},
DeBERTa, LoRA, and DeepSeek-7B achieve comparable average
F1-scores of 98.29, 98.89, and 99.17, respectively, while
exhibiting different inference, adaptation, and deployment
efficiency scores. Similarly, GPT-2 XL achieves an average F1-score of 94.77, compared with 99.17 for DeepSeek-7B, while their computational characteristics differ. These observations highlight the importance of jointly examining detection performance, inference latency, trainable parameters, and model size when assessing the practical trade-offs of LLM-based log anomaly detection.

\textbf{Impact of model quantization.} We evaluate Llama2-7B under 16-bit, 8-bit, and 4-bit precision, obtaining average F1-scores of 98.89, 99.04, and 98.98, respectively (Table~\ref{tab:performance}). Performance remains comparable across precision settings, with F1-scores ranging from 99.58 to 99.87 on BGL, 98.94 to 99.30 on HDFS, and 98.03 to 98.16 on Thunderbird. These results indicate that low-bit quantization largely preserves detection performance in the evaluated configurations, while its deployment benefits require separate assessment of computational costs.



\begin{figure}[h]
    \centering
    \includegraphics[width=0.9\linewidth]{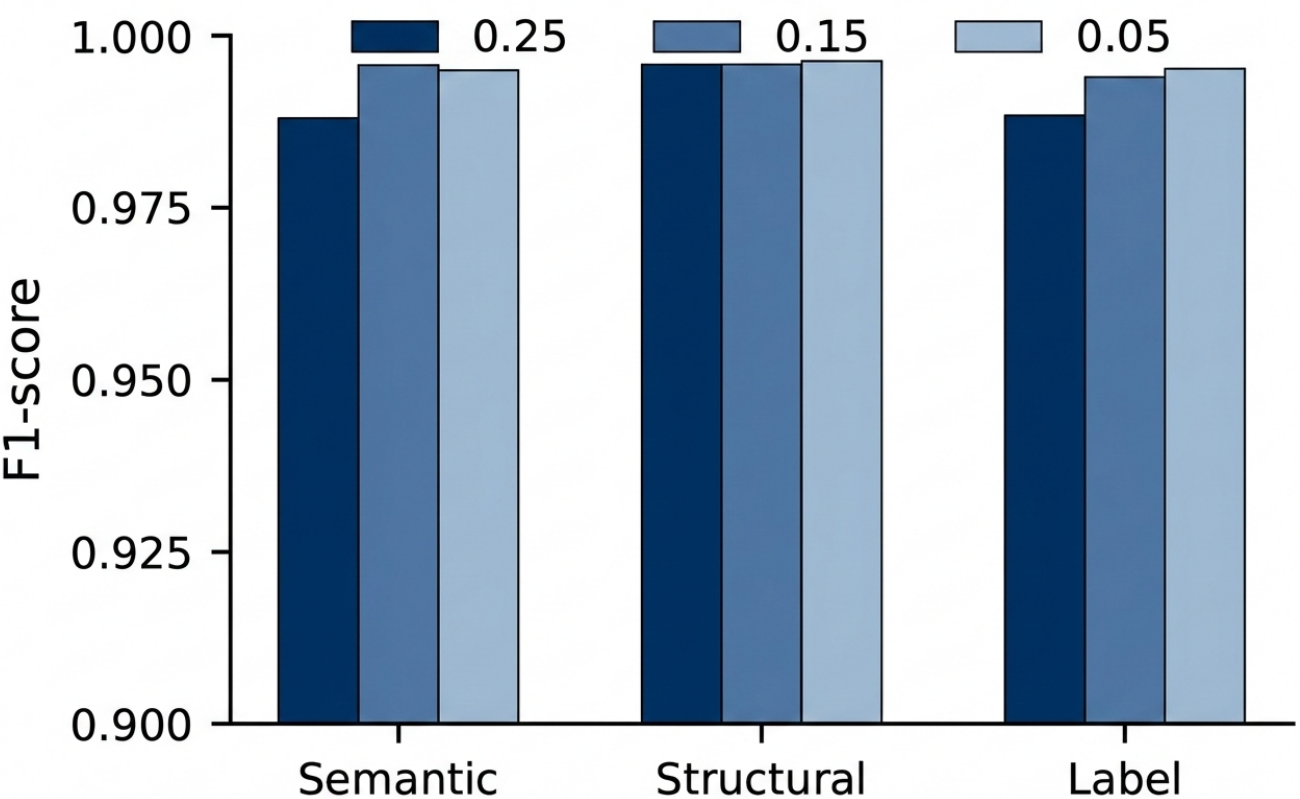}
    \caption{Detection performance of LoRA-tuned Llama2-7B under structural, semantic, and label noise at different injection ratios.}
    \label{fig:noise_eval}
\end{figure}

\textbf{Robustness under noisy logs.} We evaluate the robustness of LoRA-tuned Llama2-7B under structural, semantic, and label noise at injection ratios of 5\%, 15\%, and 25\%. As shown in Fig.~\ref{fig:noise_eval}, the model maintains high F1-scores across the evaluated settings, with structural noise causing less degradation than semantic and label noise, particularly at higher injection ratios. These results suggest greater tolerance to structural perturbations than to changes in log semantics or supervision labels.

\vspace{-2em}

\section{Conclusion}

This paper presents a systematic empirical analysis of LLM-based log anomaly detection across three public datasets. Our results reveal substantial performance differences among adaptation strategies, while the detection gains associated with model architectures and parameter scales vary across datasets. We further observe that models with comparable detection performance can exhibit different computational characteristics, and that low-bit quantization largely preserves detection accuracy in the evaluated configurations. Finally, robustness experiments examine the sensitivity of a LoRA-tuned detector to structural, semantic, and label noise. These results suggest that model selection should consider not only detection accuracy but also efficiency and robustness under practical deployment conditions. These findings highlight the importance of considering detection performance, computational efficiency, and robustness together when evaluating LLM-based log anomaly detectors.

\section{Acknowledgment}
This research was supported by the National Natural Science Foundation of China (No.62376023).

\bibliographystyle{IEEEbib}
\bibliography{strings,refs}

\end{document}